\documentclass[10pt,journal,compsoc]{IEEEtran}

\ifCLASSOPTIONcompsoc
  \usepackage[nocompress]{cite}
\else
  \usepackage{cite}
\fi

\ifCLASSINFOpdf
\else
\fi

\usepackage{amsmath,amsfonts}
\usepackage{algorithmic}
\usepackage{algorithm}
\usepackage{array}
\usepackage[caption=false,font=normalsize,labelfont=sf,textfont=sf]{subfig}
\usepackage{textcomp}
\usepackage{stfloats}
\usepackage{url}
\usepackage{verbatim}
\usepackage{graphicx}
\usepackage{cite}
\usepackage{amsmath}
\usepackage{amssymb}
\usepackage{tabu}                      
\usepackage{booktabs}                  
\usepackage{makecell}
\usepackage{multirow}
\usepackage{utfsym}
\usepackage[pagebackref=true,breaklinks=true,letterpaper=true,colorlinks,bookmarks=false]{hyperref}
\begin{document}
%
\title{JGHand: Joint-Driven Animatable Hand Avater via 3D Gaussian Splatting}
%
%
%
%

\author{Zhoutao Sun, 
	\and Xukun Shen,
	\and Yong Hu
	\and Yuyou Zhong,
	\and Xueyang Zhou 
\IEEEcompsocitemizethanks{\IEEEcompsocthanksitem The authors are with the State Key Laboratory of Virtual
	Reality Technology and Systems, Beihang University, Beijing
	100191, China (e-mail: ztsun@buaa.edu.cn, xkshen@buaa.edu.cn, huyong@buaa.edu.cn,  sy2306312@buaa.edu.cn, xyzhou97@buaa.edu.cn)\protect\\}}

\IEEEtitleabstractindextext{%
\begin{abstract}
Since hands are the primary interface in daily interactions, modeling high-quality digital human hands and rendering realistic images is a critical research problem. Furthermore, considering the requirements of interactive and rendering applications, it is essential to achieve real-time rendering and driveability of the digital model without compromising rendering quality. Thus, we propose Jointly 3D Gaussian Hand (JGHand), a novel joint-driven 3D Gaussian Splatting (3DGS)-based hand representation that renders high-fidelity hand images in real-time for various poses and characters. Distinct from existing articulated neural rendering techniques, we introduce a differentiable process for spatial transformations based on 3D key points. This process supports deformations from the canonical template to a mesh with arbitrary bone lengths and poses. Additionally, we propose a real-time shadow simulation method based on per-pixel depth to simulate self-occlusion shadows caused by finger movements. Finally, we embed the hand prior and propose an animatable 3DGS representation of the hand driven solely by 3D key points. We validate the effectiveness of each component of our approach through comprehensive ablation studies. Experimental results on public datasets demonstrate that JGHand achieves real-time rendering speeds with enhanced quality, surpassing state-of-the-art methods.
\end{abstract}

\begin{IEEEkeywords}
3D hand animation, 3D Gaussian Splatting, computer vision
\end{IEEEkeywords}}

\maketitle

\IEEEdisplaynontitleabstractindextext

%
\IEEEpeerreviewmaketitle

\IEEEraisesectionheading{\section{Introduction}\label{sec:introduction}}

%
%
%
%
\IEEEPARstart{W}{e} frequently use our hands in daily interactions, making them a crucial interface for human-computer interaction. Therefore, achieving personalized digital modeling and high-fidelity real-time rendering of hands can significantly enhance user immersion in interactive applications, making it an essential aspect of human-computer interaction and 3D computer vision research. However, significant obstacles remain in developing digital hand models that are not only easily controllable but also personalized, with real-time photorealistic rendering capability.

Previous methods\cite{moon2020deephandmesh, qian2020html, li2022nimble} use video data to learn parametric models with material maps for building personalized hand models. However, due to the limited expressive power of PCA, these models have low resolution, do not fully capture the hand shape, and the rendered images lack high-frequency details.
Leveraging the power of implicit neural rendering, some studies\cite{corona2022lisa,chen2023hand, guo2023handnerf} utilize neural radiance field(NeRF)\cite{mildenhall2021nerf} for hand rendering. These approaches treat the articulated hand as multiple rigid objects and use inverse kinematics to construct a hand model in canonical poses, thereby driving the implicit field to render images in various poses. Although NeRF-based methods can achieve arbitrary resolution rendering and produce high-fidelity images, they require extensive time for training and rendering due to their sampling and volume rendering strategies.
In order to solve this problem, a few approaches introduce a mesh-based sampling strategy\cite{mundra2023livehand, zheng2024ohta}, which significantly improves rendering speed. However, these methods require the latent shape and pose parameters from the 3D morphable model, which are more difficult for neural networks to learn and predict compared to the intuitive positions of hand joints.

The recent proposal of 3D Gaussian Splatting(3DGS)\cite{kerbl20233d} has made real-time rendering of photorealistic images possible. Research efforts have focused on extending this technique from reconstructing static scenes to dynamic objects, including human bodies\cite{qian20243dgs, shao2024splattingavatar, xiao2024neca, pang2024ash} and heads\cite{kirschstein2024gghead, xu2024gaussian}. However, the challenges posed by self-occlusion from finger movements and the richer texture details of hands prevent existing methods from being directly applied to hand modeling.

To overcome these obstacles and enable hand deformation using 3D Gaussians, we propose a novel animatable 3DGS model. Given that 3D key point coordinates are more accessible and easier for neural networks to learn, we introduce a differentiable computational process that achieves zero-error mapping from the template pose to the input pose. Utilizing this transformation and the LBS algorithm, the 3DGS can accommodate arbitrary pose and bone length changes.
Additionally, we simulate shadows caused by finger self-occlusion and produce high-quality rendered images by generating a depth image from the position and opacity of 3D Gaussians. We implement a pixel depth-based convolution kernel computation method for this shadow simulation. 
Moreover, we embed prior knowledge of the hand into the model, such as shape and the joint rotation angles, to improve the hand shape integrity and enhance the model's rendering generalization across different viewpoints and poses.

In summary, our contributions are as follows:
\begin{itemize}
	\item We propose the first joint-driven animatable 3DGS-based hand model embedded with anatomical priors, enabling real-time, photorealistic rendering of hands.
	\item We introduce a skeleton transformation which converts the canonical pose with zero-error to arbitrary poses and bone lengths. 
	\item We utilize the depth map calculated by the 3DGS and propose a real-time shadow simulation method to account for finger self-occlusion.
	\item Our extensive experiments show that our method outperforms existing state-of-the-art and prove the validity of each part of the approach.
\end{itemize}

\begin{figure*}[!t]
	\centering 
	\includegraphics[width=\linewidth]{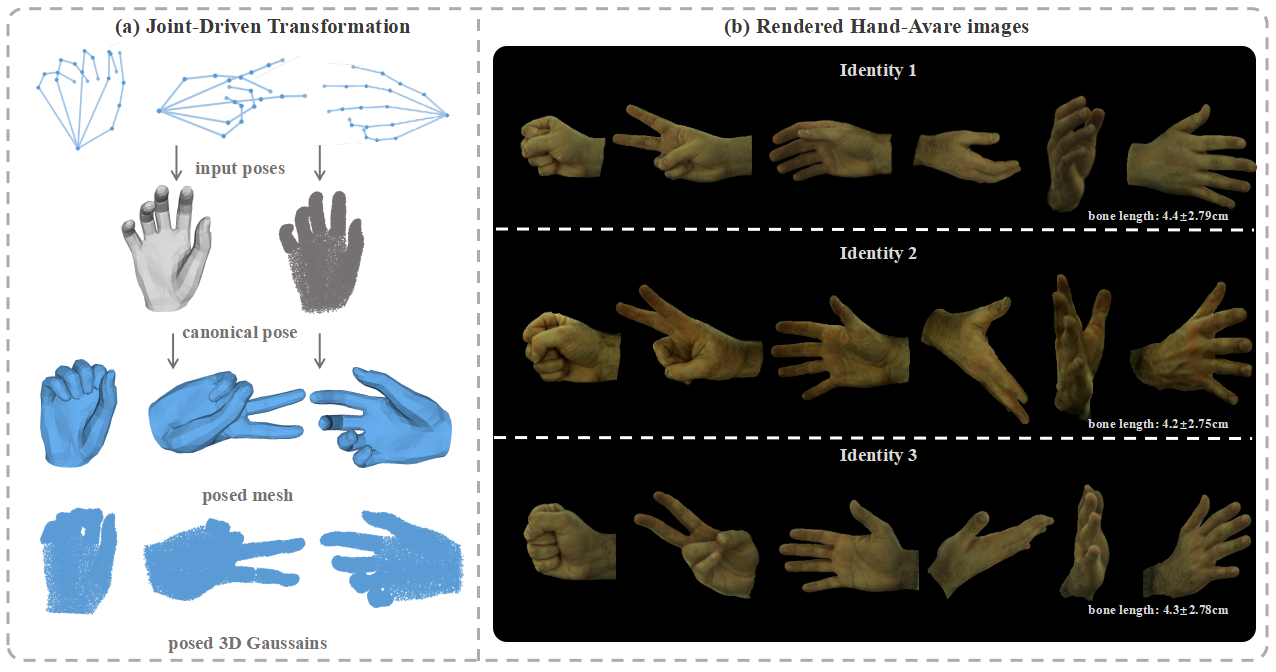}
	\caption{We present JGHand, an animatable 3DGS-based hand model driven solely by keypoints. (a) Given 3D position of hand joints, we propose a transformation that converts the canonical pose to the input pose with zero error. (b) We propose a 3DGS-based framework that reconstructs the personalized hand appearance and achieve real-time, photorealistic rendering.}
	\label{fig:main illutration}
\end{figure*}


 

\section{Related Work}
In this section, we review the most relevant existing methods for animatable hand avatar and articulated 3D Gaussians splatting. 
\subsection{Animatable Hand Avatar}
For creating personalized and animatable hand avatars, early research relied on parametric 3D morphable models that utilized low-dimensional parameters to drive the hand mesh in various shapes and poses\cite {romero2022embodied, li2022nimble}. However, these models were often too coarse, lacked texture maps, and could only recover the basic hand shape. HTML\cite{qian2020html} extends the MANO\cite{romero2022embodied} by adding a parametric hand texture model, while NIMBLE\cite{li2022nimble} developed a non-rigid parametric model simulating both bone and muscle deformations based on MRI datasets. Due to the limited vertices in the mesh and training data, these methods lack good adaptability. 

Neural rendering techniques have gained much attention for their ability to render high-quality, arbitrary-resolution images. As a result, some researchers have leveraged NeRF\cite{mildenhall2021nerf} to reconstruct animatable hand avatars. Lisa\cite{corona2022lisa} utilized articulated neural radiation fields to learn the color and geometry of hand avatars from multi-view images. HandAvater\cite{chen2023hand} employed occupation fields to recover hand geometry and estimated the albedo and illumination fields under finger self-occlusion based on volume rendering and hand geometry. However, volume rendering requires pixel-by-pixel sampling and color computation, leading to high computational complexity, long training times, and non-real-time rendering. LiveHand\cite{mundra2023livehand} and OHTA\cite{zheng2024ohta} adopted a mesh-based sampling strategy to reduce the number of sampling points and effectively constrain hand geometry. However, these approaches required accurate shape and pose parameters for the parametric 3D morphable model. It is worth noting that the above methods are based on morphable model parameters to drive the deformation of the hand avatar. Obtaining the exact parameters corresponding to the pose often requires joint image prediction or iterative optimization based on inverse kinematics. 

Karunratanakul et al.\cite{karunratanakul2021skeleton} proposed a differentiable skeleton canonicalization layer that transforms the skeleton into a canonical pose and introduced the HALO, a neural implicit surface representation of hands driven by keypoint-based skeleton articulation. However, there are two issues with HALO's transformation: it cannot accommodate changes in bone length, and it introduces errors in the transformation process. In contrast, we introduce a zero-error transformation from canonical to arbitrary pose and bone length based on a differentiable mapping computation process inspired by HALO.

\subsection{Articulated 3D Gaussians Splatting}
The 3DGS\cite{kerbl20233d} utilizes 3D Gaussians to represent static scenes and employs differentiable splatting-based rasterization for real-time, photorealistic rendering. This powerful capability has inspired several studies to extend the 3DGS to reconstruct articulated objects. Liuten et al.\cite{luiten2023dynamic} first proposed using time as an attribute of 3D Gaussians for dynamic human representation, enabling the reconstruction of different human body poses within a sequence segment. However, this method can only render high-precision images of human poses appearing within the sequence from random viewpoints and cannot be extended to arbitrary poses. 

A natural idea is to reconstruct a canonical pose template and utilize Linear Blend Skinning(LBS) to drive the 3DGS-based human avatars\cite{hu2024gauhuman, kocabas2024hugs, qian20243dgs, lei2024gart}. These methods offer solutions for building animatable human avatars via 3DGS and focus on overcoming obstacles in the field of human reconstruction, such as handling clothing and reducing artifacts. Li et al.\cite{li2024animatable} and Liu et al.\cite{liu2024gea} utilized the parameterized SMPL-X model\cite{pavlakos2019expressive} to establish 3DGS-hased human avatars that included hands. 
However, compared to the human body, finger movements cause more severe self-occlusion, making it more difficult to recover the full hand shape. Additionally, the flexible movements of fingers create shadows, complicating the recovery of high-frequency textures.
Pokhariya et al.\cite{pokhariya2024manus} proposed the MANUS, a method for hand representation using 3DGS. Nevertheless, the MANUS focused on estimating hand-object contact and recovering accurate hand geometry, resulting in rendered images with poor realism. In contrast to these approaches, we propose an animatable hand avatar using 3DGS. By modeling reasonable shadows, our model can render high-fidelity hand images in real time with arbitrary poses.

\begin{figure*}[htb]
	\centering 
	\includegraphics[width=\linewidth]{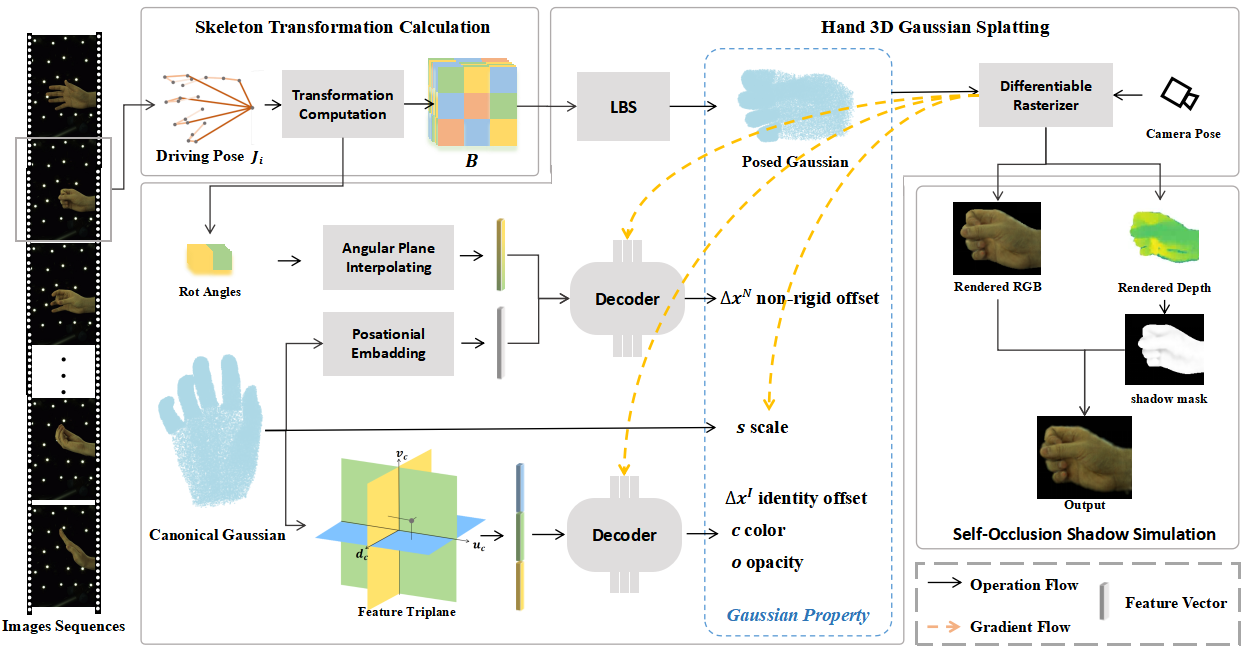}
	\caption{An overview of our proposed framework. Given a hand pose and a camera view from an RGB sequence, our method reconstructs an identity hand avatar and renders a photorealistic hand image in real-time. First, we compute the transformation based on the given hand pose. The estimation of the 3D Gaussian attributes is performed using the UVD coordinates of the canonical Gaussian. In this process, the position of the Gaussian is calculated using the transformation and the LBS algorithm. During the hand image rendering, the depth value of each pixel is computed, followed by the simulation of self-occluding shadows. The rendered image and simulated shadows are then superimposed to produce the final output.}
	\label{fig:pipline}
\end{figure*}

\section{Methods}
Given a sequence of multi-view or single-view RGB hand images of a subject, $\{G_i \mid i = 1, \ldots, N\}$ for $N$ frames, our method generates an animatable 3DGS-based hand model capable of rendering photorealistic hand images in real time. \autoref{fig:pipline} provides an overview of our proposed framework. First, we compute the skeleton transformation $B$ based on the hand pose $J_i$ in $G_i$(see Section \ref{section: Skeleton Tranformation Calculation}). For the canonical Gaussians driving and image rendering of input pose(see Section \ref{section: Hand 3D Gaussian Splatting}), we proceed using Linear Blend Skinning (LBS) and 3D Gaussian Splatting. Finally, to address shadows generated by the self-occlusion of finger movements, we introduce a depth-based shadow simulation method (see Section \ref{section: Self-Occlusion Shadow Simulation}) and combine it with 3DGS-rendered images to produce high-fidelity results.

\subsection{Skeleton Transformation Calculation}
\label{section: Skeleton Tranformation Calculation}
A hand skeleton $J \in \mathbb{R}^{21 \times 3}$, expressed in keypoint coordinates where 21 denotes the number of hand joints, is used to obtain the transformation matrix, $B=\{B_i\mid i=1,\ldots,21\}$, where $B_i \in SE(3)$, from the $J^c$ to $J$. The $J^c$ represents the joint locations of the canonical gaussian, and its acquisition is described in Section \ref{section: Hand 3D Gaussian Splatting}. Formally,
\begin{equation}
	J = BJ^c
\end{equation}

\begin{figure}[tb]
	\centering 
	\includegraphics[width=\linewidth]{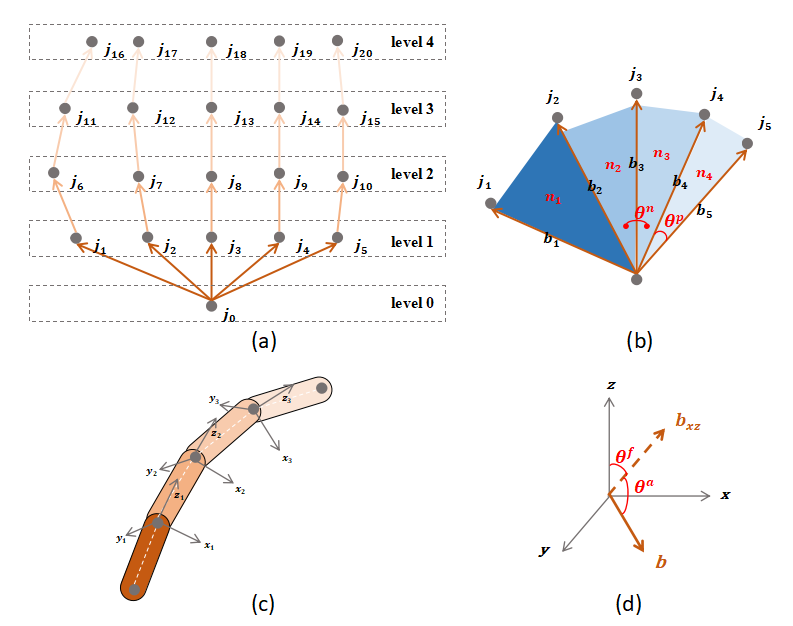}
	\caption{(a) is an illustration of hand joints and levels, and the node with level 0 is the root joint. 
		(b) illustrates the planes defined by the root joint and the level 1 joints.
		(c) indicates the local coordinate systems for each joint point on one finger.
		(d) shows the rotation angles of a joint in the local coordinate systems. $b_{xz}$ is the projection of the bone vector $b$ onto the $xz$ plane. The abduction angle $\theta_a$ is the angle between $b$ and $b_{xz}$, and the flexion angle $\theta_f$ is the angle between $b_{xz}$ and the coordinate axis $z$.}
	\label{fig:handjoints}
\end{figure}

\textbf{Notation}. As shown in \autoref{fig:handjoints}, $J = \{j_i \mid i=1,\ldots,20\}$ are the coordinates of $21$ hand joints, and $\{b_i \mid i=1,\ldots,20\}$ represent the bone vectors between the $i^{th}$ joint and its parent joint. For the input skeleton, the pose is defined by a set of angles computed from the joints' positions.
For bones in the palm plane (connected by level-1 and the root joint), $\{n_i \mid i = 1, \ldots, 4\}$ denotes the plane defined by the neighboring bones. $\theta^p_{i,j}$ is the angle between the bones $b_i$ and $b_j$. $\theta^n_{i,j}$ is the angle between the neighboring planes $n_i$ and $n_j$.
For each non-zero level joint, $\theta^a$ and $\theta^f$ denote the abduction and flexion angles between its connecting bones, respectively. In the subsequent description, variables with superscript $c$ refer to those in the canonical pose.

In general, the purpose of the transformation $B$ is to adjust the angles mentioned above and the bone lengths in $J^c$ to match those in $J$. There are several steps to convert the canonical pose $J^c$ to the target pose $J$. The steps are as follows:
1) Mapping to unit bone vectors. Convert $J^c$ to unit length bone vectors.
2) Transforming to local systems. Define local coordinate systems based on the parent joint of each bone, convert all the bone vectors to the local coordinates, and compute the rotation angles.
3) Rotating in local systems and mapping back to global system. 
Based on the non-zero hierarchical rotation angles $\{\theta^{a,f}_i \mid i = 1, \ldots, 20\}$, rotate the bone vectors based on the kinematics hierarchy to be consistent with the target pose.
4) Scale to bone lengths and mapping back. Restore the rotated bone vectors to the target bone length, align them to their parent joints, and restore the joint points.
5) Aligning the palm plane.
Rotate the joint points finger by finger according to the hand plane angles $\theta^p, \theta^n $ to achieve the target pose. Formally,

\begin{equation}
	B = PK'F'RFK
\end{equation}
where $K$ is a matrix maps the $J^c$ to bone vectors of unit length originating from the origin, respectively. $F$ is a function that transforms these bone vectors into the defined local systems. $R$ is the rotatioin matrix that rotates the vectors by specified angles within the local coordinate systems. $F'$ maps the rotated vectors back to the global coordinate system, following the kinematics hierarchy. $K'$ scales the vectors to the bone lengths of $J$ and  maps them back to the coordinates. $P$ denotes the rotation matrix associated with the angles of the palm planes.

The definition of the local coordinate systems transformation $F$ and the rotation matrix $R$ are consistent with HALO, more detailed information can be found in \cite{karunratanakul2021skeleton}. 
To address bone length and rotation angle errors during the conversion, we detail here only the computational steps that differ from HALO. For any non-root joint $j_i$, its corresponding matrix $K_i \in R^{4 \times 4}$ is defined as followed:
\begin{equation}
	K_i = 
	\begin{bmatrix}
		\frac{1}{||b_i^c||}I(3) & -j_{p(i)} \\
		0 & 1
	\end{bmatrix}
\end{equation}
where $I(3) \in R^{3 \times 3}$ is a diagonal matrix and $j_{p(i)}$ denotes the parent joints of $j_i$. $||b_i^c||$ represents the bone length of $b_i$ in the canonical pose. 

Since the rotation of the parent bone vector affects the $i^{th}$ bone, $F'_i \neq F^{-1}_i$. First define the global rotation $G_i$ of the local system according to the kinematics chain. The final transformation $F'_i$, which maps the local system to the global system, can be calculated as:
\begin{equation}
	\begin{aligned}
		&G_i = \begin{cases}
			F_{p(i)} R_{p(i)} F^{-1}_{p(i)} & \text{if } i \neq 1,2,3,4,5\\
			I(4) & \text{if } otherwise
		\end{cases}	\\
		&F'_i = G_{p(i)} F^{-1}_i
	\end{aligned}
\end{equation}

Similar to the process of calculating of $F^{-1}_i$, the calculation of $K'_i$ also requires following the kinematic chain. The unit bone vectors need to be rotated to the posed direction in the global coordinate system and then be scaled.
\begin{equation}
	\begin{aligned}
		b_i^{posed} &= F'_i R_i F_i b_i^c \cdot ||b_i|| \\
		t_i &= \begin{cases}
			b_{p(i)}^{posed} + t_{p(i)} & \text{if } i \neq 0\\
			0& \text{if } otherwise
		\end{cases}	\\
	\end{aligned}
\end{equation}
$||b_i||$ is the bone length of $b_i$ in the target pose $J$. $b_i^c$ denotes the unit length canonical bone vector. $F'_i, R_i,$ and $F_i$ represent the $i^{th}$ elements in the transformation matrices $F', R,$ and $F$, respectively. After obtaining the bone vectors with the target length, $K'_i$ transfers the $b_{p(i)}^{posed}$ to the tip of the parent bones based on the kinematic hierarchy.
\begin{equation}
	K'_i = 
	\begin{bmatrix}
		I(3) & t_i \\
		0 & 1
	\end{bmatrix} \\
\end{equation}

We have demonstrated that our method, compared to HALO\cite{karunratanakul2021skeleton}, can convert the canonical pose to arbitrary bone lengths and eliminate errors before and after transformation (see Section \ref{section: Comparison of Transformation}).

\subsection{Hand 3D Gaussian Splatting}
\label{section: Hand 3D Gaussian Splatting}

To render hand images from the canonical Gaussian and input poses, our method can be systematically divided into the following steps: First, we transform the Gaussian from the canonical space to the posed space. Second, we estimate the properties of the 3D Gaussian using identity-specific, learnable features. This structured approach ensures accurate rendering by dynamically adapting the Gaussian's attributes according to the pose and personalized characteristics.

\begin{figure}[tb]
	\centering 
	\includegraphics[width=0.9\linewidth]{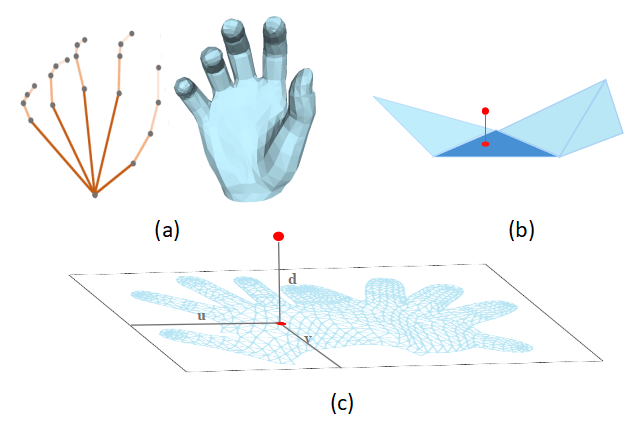}
	\caption{(a) represents the joint positions and mesh of MANO with the mean pose and shape parameter. (b) shows a sampling point located inside the canonical pose mesh, along with the nearest mesh face to it. The diagram includes four triangles that represent partial facets of the mesh. The upper red points indicate the sampling points, while the lower ones mark the projection points. The dark blue triangle highlights the facet where the projection points are located. (c) is an illustration of the uvd coordinates of the sampling point.}
	\label{fig:uvd coord}
\end{figure}

\textbf{Canoncial Gaussian transformation}.
Inspired by the strategy of mesh-based sampling in Livehand\cite{mundra2023livehand}, our method utilize a 3d Gaussian-based template with canonical pose to ensure the integrity of the generated hand shape. 
The canoncial Gaussians are initialized based on the MANO\cite{romero2022embodied} model with the mean pose and shape parameter, as shown in \autoref{fig:uvd coord}a. Initially, we perform random sampling within the mesh, categorizing the sample points according to their proximity to the nearest bone of the canonical skeleton. This sampling process is repeated until the predetermined number $N_b$ of sample points for each bone is achieved. The spatial coordinated of these sampled points serve as the initial positions for the canonical Gaussians. 

For a more efficient representation, we utilize the uv map of the parameterized model to obtain the normalized coordinates $(u,v,d) \in [0,1]^3$ of each Gaussian, as shown in \autoref{fig:uvd coord}, enhancing 3D Gaussian property estimation. We calculate the $(u,v)$ coordinates of each sampling point by determining the projection point on the nearest face of the parametric mesh and employing barycentric coordinate interpolation. The initial $d$-coordinate is determined by measuring the distance from the sampling point to its projection point. Subsequently, we normalize the $d$-coordinates of all sampling points to ensure consistent scaling across the model.

To obtain the positions of the Gaussians in the posed space, we employ forward kinematics. Traditional LBS is applied to 2D mesh surfaces, leading previous methods to rely on the nearest vertices of the parameterized model to extend skinning to 3D. This approach, however, often results in spatial discontinuities\cite{bhatnagar2020loopreg} or necessitates optimization via neural networks\cite{qian20243dgs,kocabas2024hugs}. 
In contrast, our method leverages Fast-SNARF\cite{chen2023fast} to create a weight field. Utilizing the positions of the Gaussians, we can interpolate within this field to derive the skinning weights $W$, enhancing the spatial continuity and accuracy of the deformation. Given the $i^{th}$ Gaussian position $p_i^c\in R^3$ in the canonical space, the posed Gaussian position $p_i^{posed}=(\sum_{j=1}^{21}W_jB_j) p_i^c$, where $W_j$ and $B_j$ denote the item in $W$ and $B$ corresponding to the $j^{th}$ joint.

\textbf{3D Gaussian property estimation}.
The 3DGS\cite{kerbl20233d} represents a static scene using a set of 3D Gaussians. Each Gaussians is characterized by several parameters: 3D center position $x \in R^3$, spherical harmonic(SH) coefficients for colors computation from various directions, opacity $o \in R$, 3D rotation $q \in SO(3)$, scaling factors $s \in R^3_+$ along the Gaussian axes. Given these properties, The 3DGS can render images via a differentiable rasterization process. 
During the training process, the parameters of the 3D Gaussian are optimized and operations such as pruning and copying are performed. These modifications are strategically implemented to progressively align the rendered image with the training image, enhancing the visual similarity between them. Different from the original 3DGS, we have implemented several modifications in the proposed approach.

Due to the sparse viewpoints in the training data, we adjust the the 3D Gaussian from anisotropic to isotropic. Instead of optimizing the SH coefficients, we now directly predict RGB values. Furthermore, we unify the scaling across all three dimensions, and the rotation $q$ is fixed at [1,0,0,0]. In summary, our modifications transform the original 3D Gaussian into a sphere characterized by a fixed color and arbitrary size, better suited to our dataset constraints.

Considering the creation of personalized hand shapes and maintain consistency in textures, our method employs a trainable identity feature triplane. We perform interpolation within these triplanes using the normalized $uvd$-coordinates of the canonical Gaussian to obtain the feature vector. The vector is then processed using MLPs to predict the Gaussian's identity-specific offset $\triangle x^I$, color $c$, and opacity $o$.
Additionally, we consider the non-rigid deformation of the hand in different pose and propose a pose-aware non-rigid offset $\triangle x^N$. The feature vector utilized for non-rigid offset prediction is concatenated by a position vector and the angular feature. The position vector is obtained from the $uvd$-coordinates by positional encoding from Nerf\cite{mildenhall2021nerf}:
\begin{equation}
	\lambda(u) = (sin(2^0\pi u), cos(2^0\pi u), \ldots, sin(2^{L-1}\pi u), cos(2^{L-1}\pi u))
\end{equation}
the function $\lambda$ is applied separately to each of the three coordinates values in $uvd$-coordinates, and $L$ is the specified dimension. The angular features are derived by interpolating in trainable feature planes, using normalized joint rotation angles—specifically, abduction and flexion angles obtained from transformation computations. We normalize these angles by calculating their extremes, referencing methodologies from \cite{chen2013constraint, spurr2020weakly}. Furthermore, to maintain the influence of the parent joint, the final angular features are hierarchically encoded according to the kinematic tree.
\begin{equation}
	F_i = \begin{cases}
		\frac{1}{2}[F_{p(i)} + \delta(\theta_{i})] & \text{if } i \neq 1,2,3,4,5\\
		\delta(\theta_{i})& \text{if } otherwise
	\end{cases}	\\
\end{equation}
$\theta_{i}$ indicates the normalized rotation angles and the function $\delta(\cdot)$ refers to the interpolation from the feature planes. The index $p(i)$ denotes the parent joint of $i^{th}$ joint.

To ensure the accurate influence of pose-aware and identity-aware offsets on the coordinates of the 3D Gaussians, we compute the positions of the $i^{th}$ posed Gaussian for the rasterizer, denoted by $x_i$, as follows:
\begin{equation}
	x_i = (\sum_{j=1}^{21}W_jB_j) (p^c_i + \triangle x^I_i) + \triangle x^N_i
\end{equation}

Moreover, to prevent the offsets from becoming excessively large and disrupting the hand's shape, our method incorporates a position regularization during training. 
\begin{equation}
	L_{reg}^{offset} = ||\triangle x||_2
\end{equation}

This approach encourages the Gaussian positions to remain close to the canonical template, ensuring a more stable and accurate representation of the hand's structure.

\subsection{Self-Occlusion Shadow Simulation}
\label{section: Self-Occlusion Shadow Simulation}
\begin{figure}[tb]
	\centering 
	\includegraphics[width=0.9\linewidth]{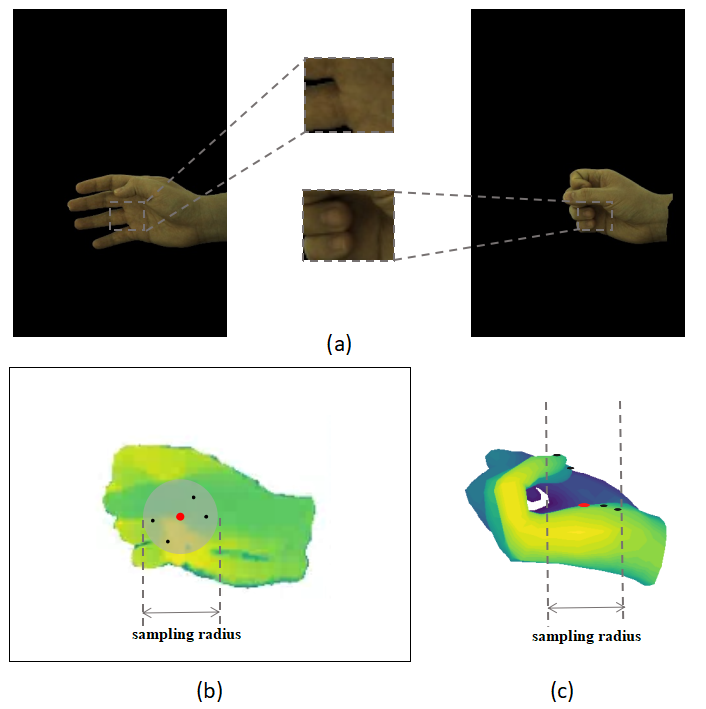}
	\caption{The top row of images demonstrates the varying shadow effects in the palm area resulting from finger movements, while the bottom row visualizes a pixel depth-based convolution kernel used to process these shadows. The red point in (b) marks a pixel for which a shadow mask will be calculated, the gray area delineates the region to be sampled, and black points identifies several specific sampling point within this region. (c) presents a side view of a mesh that maintains the same pose as seen in (b), and it maps the points from (b) directly onto that mesh surface.}
	\label{fig:shadow cal}
\end{figure}

As illustrated in the \autoref{fig:shadow cal}a, varying hand poses create distinct patterns of light and shadow across the hand. A natural idea to address this obstacle is to develop a pose-aware shadow estimation neural network, but this approach requires extensive training data and lacks robustness. Drawing inspiration from Screen Space Ambient Occlusion (SSAO)\cite{bavoil2008screen}, we propose a differentiable shadow calculation layer designed to simulate self-occlusion shadows caused by finger movements, enhancing the visual realism of the rendered image.

During rendering, we calculate the depth image of 3D Gaussian simultaneously. The depth of a pixel $d$ is determined by opacity blending of the $N$ contributing Gaussians, which are sorted from nearest to farthest:
\begin{equation}
	d = \sum_{j=1}^{N}d_j o_j \prod_{k=1}^{j-1}(1 - o_k)
\end{equation}
where the $d_j$ is the depth of the Gaussian. we generate a convolution kernel based on the specified sampling radius and the number of samples. Each element within the kernel represents the offset from the sampling point to the pixel, as depicted in \autoref{fig:shadow cal}b. This convolution kernel is subsequently applied across all pixels of the depth image to produce a shadow mask, denoted as $S$:
\begin{equation}
	S_{x} = \frac{1}{N}\sum_{i=1}^{N}[f(d(x), d(x+\triangle x_k))]
\end{equation}
where $x$ represents the pixel coordinates and $N$ represents the total number of sampling points. The term $\triangle x$ specifies the offsets from the pixel to the respective sampling points, while $d(\cdot)$ refers to a function that maps to the depth value. Additionally, $f(a,b)$ is defined as a differentiable mapping function that returns values between 0 and 1, used to quantify the relative magnitude of two elements. 

At this stage, we obtain an RGB image through differentiable rasterization and a shadow mask from the rendered depth image. These two are then combined pixel by pixel to produce the final rendered image.

\subsection{Optimization}
We optimize the parameters of the two decoders, the angular plane, the feature triplane, and the scales of the Gaussians. We compare the rendered image with the ground-truth image and the hand segmentation mask for loss function calculation. Specifically, our loss is composed of:
\begin{equation}
	\begin{aligned}
		L &= \lambda_{rgb} L_{rgb} + \lambda_{ssim} L_{ssim} + \lambda_{lpips} L_{lpips} + \lambda_{mask} L_{mask} \\
		&+ \lambda_{reg} L_{reg} +
		\lambda_ + \lambda_{iso} L_{iso} 
	\end{aligned}
\end{equation}
where $L_{rgb}$, $L_{ssim}$ and $L_{lpips}$ are the L1 loss, SSIM loss\cite{wang2004image} and LPIPS loss\cite{zhang2018unreasonable}, respectively. $L_{mask}$ corresponds the L2 loss calculated between the rendered hand region and the ground-truth mask. Additionally, $L_{reg}$ is the regularization terms applied to the offset. $L_{iso}$ is an isotropic regularization, adopted from \cite{pokhariya2024manus}, which ensures that the optimized Gaussians remain as isotropic as possible. We set $\lambda_{rgb}=1$, $\lambda_{ssim}=0.2$, $\lambda_{lpips}=0.2$, $\lambda_{mask}=0.2$, $\lambda_{reg}=1$ and $\lambda_{iso}=0.05$.

\section{Experiments}
\textbf{Datasets}. We compare our method with the SOTA methods on two datasets. The InterHand2.6M\cite{moon2020interhand2} dataset is a large-scale,  real-captured dataset with multi-view sequences, containing various hand pose from 26 unique subjects. For a fair comparison, we choose `test/Capture0', `test/Capture1' and `val/Capture0' from the dataset for training and validation, following\ the approach in \cite{chen2023hand}. The HandCo\cite{zimmermann2021contrastive} dataset is a synthetic dataset containing hand images captured by 8 cameras with background replacement. We utilize the sequence `0191' as describe in \cite{zheng2024ohta}.

\textbf{Implementation Details}. Our method is implemented on a single NVIDIA RTX 3090 with the PyTorch\cite{paszke2019pytorch} framework. The number $N_b$ of sample points of each bone in the canonical Gaussian is $3000$, and the number of convolution kernel sampling points in shadow simulation is set to $64$. Additionally, the decoders consist of three MLPs. The network is trained by iterating 30 epochs and the learning rete is set to $1e-3$.

\textbf{Metrics}.
In our experiments, we use the Mean Per Joint Position Error (MPJPE) and Chamfer-distance(L1) to measure the correctness of the transformation.
Additionally, we utilize LPIPS, PSNR, and SSIM to measure image similarity as metrics of rendering quality.

\subsection{Evaluating Skeleton Transformation} 
\label{section: Comparison of Transformation}
Since the computational process of our skeleton transformation is based on HALO \cite{karunratanakul2021skeleton}, we constructed our experiments by comparing our results with theirs. Additionally, because the transformation in HALO maps the skeleton from posed space to canonical space, we utilize its irreversible transformation.

\begin{figure}[tb]
	\centering 
	\includegraphics[width=\linewidth]{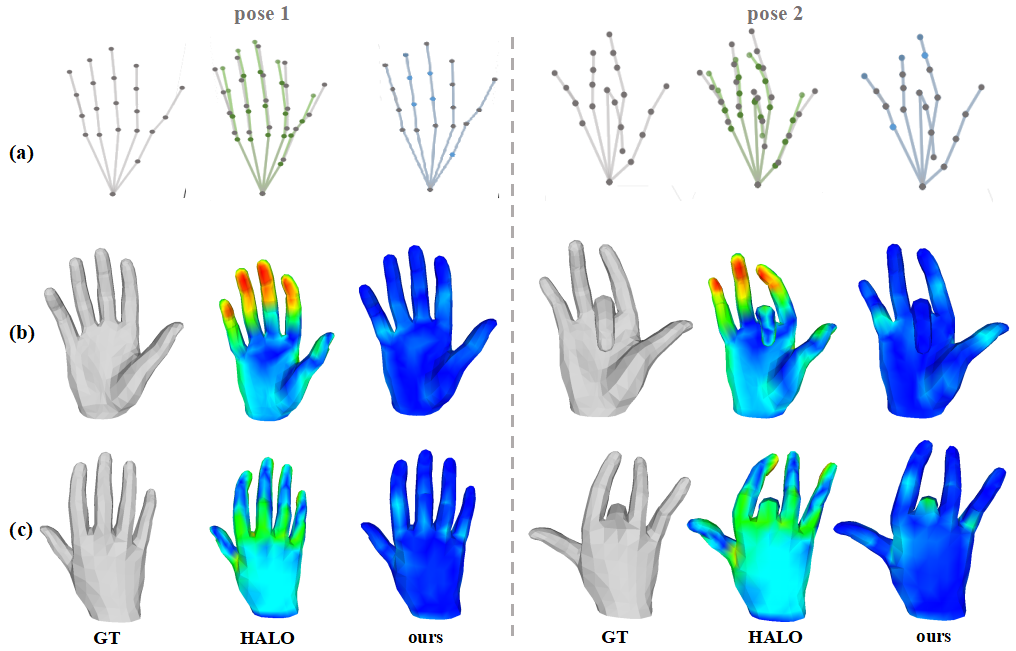}
	\caption{The top row shows the joints and bones. Grey points and lines represent the ground truth, while green points indicate positions mapped from the canonical pose using the HALO transformation\cite{karunratanakul2021skeleton}. The blue points represent positions transformed using our method. The bottom two rows display the front and back views of the converted mesh. Vertex colors denote the Hausdorff Distance between the transformed mesh and the ground truth: bluer colors indicate smaller distances, and redder colors indicate larger distances.}
	\label{fig:transform compare}
\end{figure}

We utilize the skeleton and mesh from MANO\cite{romero2022embodied} with the mean pose and shape parameters as the canonical space. Based on the input skeleton, we compute the transformation matrix and transform the canonical template to the posed space. We utilize the transformed joint coordinates and vertex positions to compare with the ground-truth, and the qualitative comparison is shown in \autoref{tab:transformation compare}. We randomly select two skeletons, and the qualitative evaluation is illustrated in \autoref{fig:transform compare}. Compared to HALO, our method can convert the joint positions in the canonical space to the input pose without error and also reduces the error of 3D LBS by 83\%. The only remaining error in 3D LBS is mainly since our transformation relationship can only realize the variation of bone length, which cannot reflect some personalized shape changes of the hand.

\begin{table}[htb]
	\caption{Comparison between HALO\cite{romero2022embodied} and our transformation.  $\downarrow$ means the lower the better. }

	\label{tab:transformation compare}
	\small%
	\centering%
	\begin{tabular}{%
			l|
			c c |c c%
		}
		\toprule
		\multirow{2}*{Method} &\multicolumn{2}{c|}{test/capture0}& \multicolumn{2}{c}{test/capture1} \\ \cline{2-5}
		\rule{0pt}{10pt} & MPJPE$\downarrow$ & Cham.$\downarrow$&
		MPJPE$\downarrow$ & Cham.$\downarrow$\\
		\midrule
		HALO\cite{romero2022embodied} & 0.0103& 5.14&0.0120&5.97\\
		ours & \textbf{0} & \textbf{0.92}&\textbf{0}&\textbf{0.97}\\
		
		\bottomrule
	\end{tabular}%
\end{table}

\subsection{Evaluating Rendering quality} 
\begin{figure}[tb]
	\centering 
	\includegraphics[width=\linewidth]{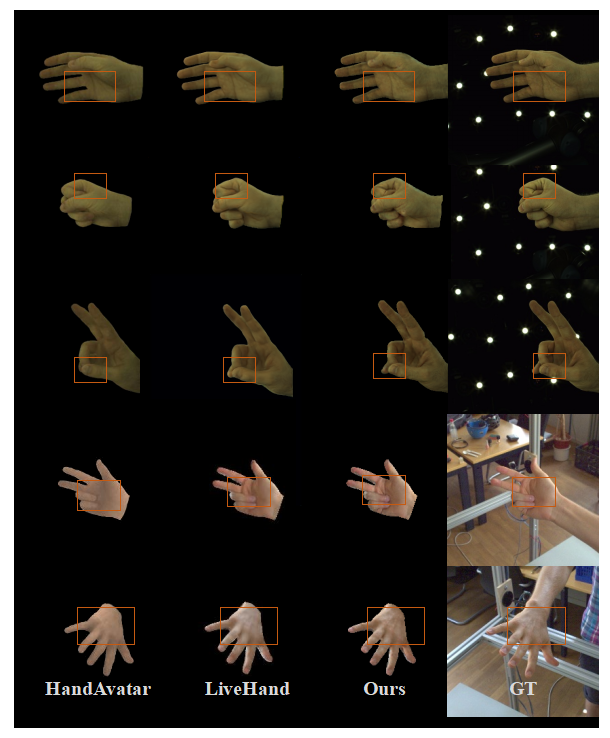}
	\caption{Qualitative results of previous methods and our method on different hand poses. }
	\label{fig:compare SOTA}
\end{figure}

To ensure a fair comparison of hand avatar reconstruction and rendering quality with previous methods, we re-trained them on the aforementioned datasets.
Since 3D-PSHR\cite{jiang20233d} has not released code but uses consistent datasets with us, we reference their reported results. The quantitative analysis is presented in \autoref{tab:render result}. Notably, while all comparison methods rely on shape and pose parameters of the 3D morphable model to drive the hand avatar, our approach only requires the more intuitively available joint coordinates for transformation. Additionally, since HandAvatar\cite{chen2023hand} and LiveHand\cite{mundra2023livehand} utilize ray sampling for rendering, whereas our method leverages the real-time rendering capabilities of 3DGS\cite{kerbl20233d}, we achieve not only improved rendering quality but also significantly faster rendering speed, as demonstrated in \autoref{tab:inference time}.
Rendering results are illustrated in \autoref{fig:compare SOTA}, showing that our method enhances rendering quality, recovers more detailed textures, and reduces rendering time.
\begin{table*}[htb]
	\caption{Comparison of rendering images on different datasets between different methods. }
	\label{tab:render result}
	\small%
	\centering%
	\begin{tabular}{%
			l |c c c| c c c| c c c | c c c%
		}
		\toprule
		\multirow{2}*{Method}& \multicolumn{3}{c|}{test/capture0}& \multicolumn{3}{c|}{test/capture1}&
		\multicolumn{3}{c|}{val/capture0} &
		\multicolumn{3}{c}{HandCo/191}\\ \cline{2-13}
		\rule{0pt}{10pt} & SSIM$\uparrow$ & PSNR$\uparrow$ & LPIPS$\downarrow$ & 
		SSIM$\uparrow$ & PSNR$\uparrow$ & LPIPS$\downarrow$ & 
		SSIM$\uparrow$ & PSNR$\uparrow$ & LPIPS$\downarrow$&
		SSIM$\uparrow$ & PSNR$\uparrow$ & LPIPS$\downarrow$ \\
		\hline
		\rule{-3pt}{10pt} HTML\cite{qian2020html} &
		0.859 & 24.23 & 0.181 &
		0.853 & 23.11 & 0.173 & 
		0.851 & 23.41 & 0.186 &
		- & - & - \\
		
		3D-PSHR\cite{jiang20233d} &
		0.934 & 30.93 & 0.078 & 
		0.913 & 29.23 & 0.089 & 
		0.910 & 29.40 & 0.092 &
		- & - & - \\ 
		
		HandAvater\cite{chen2023hand} &
		0.954 & 30.93 & 0.042 &
		0.947 & 28.80 & 0.052 & 
		0.954 & 30.28 & 0.041 & 
		0.953 & 29.81 & 0.039\\
		
		LiveHand\cite{mundra2023livehand} &
		0.960 & 32.32 & \textbf{0.032} & 
		0.959 & 31.47 & 0.033 & 
		0.786 & 29.91 & 0.043 & 
		0.967 & 31.08 & 0.028\\
		
		our & \textbf{0.966} & \textbf{33.44} & \textbf{0.032} &
		\textbf{0.966} & \textbf{32.44} & \textbf{0.032} & 
		\textbf{0.969} & \textbf{33.41} & \textbf{0.031} &
		\textbf{0.974} & \textbf{33.89} & \textbf{0.018} \\
		\bottomrule
	\end{tabular}%
\end{table*}

\begin{table}[htb]
	\caption{Average inference time to render a image on the test set.}
	\label{tab:inference time}
	\small%
	\centering%
	\begin{tabular}{%
			c|c|c|c
		}
		\toprule
		Method & HandAvater\cite{chen2023hand} &
		LiveHand\cite{mundra2023livehand} &
		ours\\
		\midrule
		Time(s)$\downarrow$ & 4.317 & 0.087 & 0.040\\
		
		\bottomrule
	\end{tabular}%
\end{table}

\subsection{Ablation Study}
\begin{figure}[tb]
	\centering 
	\includegraphics[width=\linewidth]{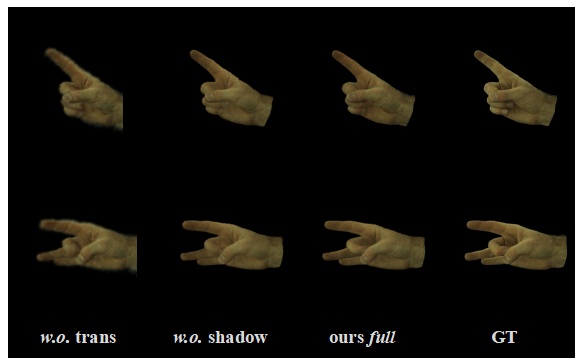}
	\caption{ Ablation study illustrating the visualization of rendered hand images from different driving-poses.}
	\label{fig:ablation study}
\end{figure}

\begin{figure}[tb]
	\centering 
	\includegraphics[width=\linewidth]{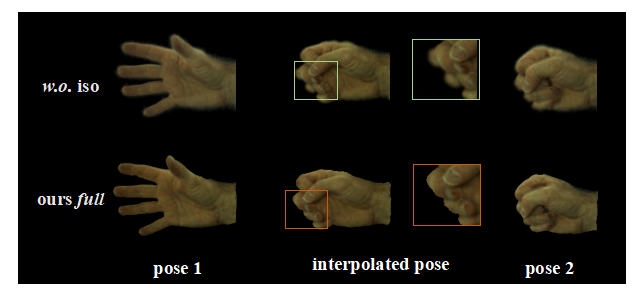}
	\caption{ Ablation study showing the rendering results from interpolated poses based on the proposed method with anisotropic and isotropic Gaussian.}
	\label{fig:wo iso}
\end{figure}

We conduct the ablation study on the 'test/capture0' sequence from InterHand2.6M\cite{moon2020interhand2}. The quantitative results are shown in \autoref{tab:ablation study} and the visualizations are illustrated in \autoref{fig:ablation study} and \autoref{fig:wo iso}. We first demonstrate the impact of accurate transformation(`$w.o.$ trans.') on 3D Gaussian reconstruction for articulated hands. Errors in transformation can lead to discrepancies between the posed Gaussian positions and the rendered image, particularly at the fingertips. This results in inaccuracies in learning the correct attributes of the 3D Gaussians in error-prone regions. Additionally, we verify the effectiveness of our proposed shading simulation strategy(`$w.o.$ shadow'). Both numerical and visual results show that simulating shadows significantly enhances the realism of the rendered images. Furthermore, we compare the effects of using isotropic versus anisotropic Gaussians (`$w.o.$ iso'). According to \autoref{tab:ablation study}, anisotropic Gaussians yield a slightly higher numerical index. However, their stability is poorer. As shown in \autoref{fig:wo iso}, when generating a novel driving pose based on two input skeletons and their joint rotation angles, the rendering with anisotropic Gaussian exhibits greater stability in response to the new pose compared to isotropic Gaussian.

\begin{table}[htb]
	\caption{Ablation study on different components from proposed method. }
	\label{tab:ablation study}
	\small%
	\centering%
	\begin{tabular}{%
			l|c c c%
		}
		\toprule
		\rule{0pt}{10pt} & SSIM$\uparrow$ & PSNR$\uparrow$ & LPIPS$\downarrow$\\
		\hline
		\rule{-3pt}{10pt} $w.o.$ trans. &0.923 &27.04 &0.097 \\
		
		$w.o.$ shadow &0.954 & 30.92&0.043 \\
		
		$w.o.$ iso.  & \textbf{0.966} & \textbf{33.56} & \textbf{0.032} \\
		
		\textbf{ours} $full$& \textbf{0.966} & 33.44 & \textbf{0.032}\\
		\bottomrule
	\end{tabular}%
\end{table}

\subsection{Novel Animation Rendering Results}
\begin{figure*}[!t]
	\centering 
	\includegraphics[width=0.9\linewidth]{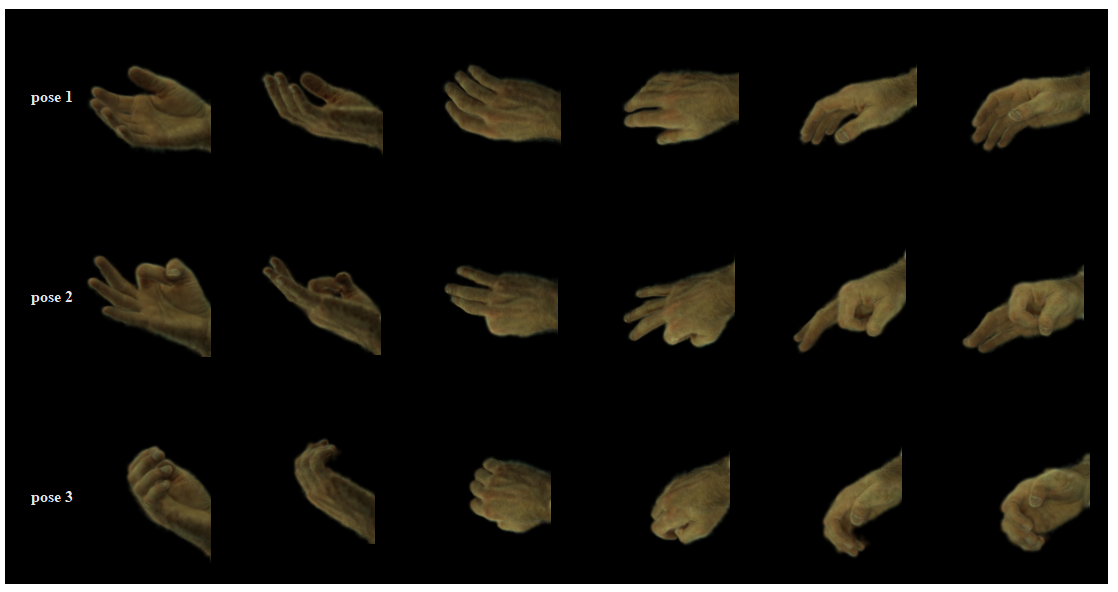}
	\caption{Qualitative results of three different driving poses in novel viewpoints. Each row illustrates, from left to right, rendered images of hand avatar driven by the same skeleton from various camera views.}
	\label{fig:novel view}
\end{figure*}

\begin{figure*}[!t]
	\centering 
	\includegraphics[width=0.94\linewidth]{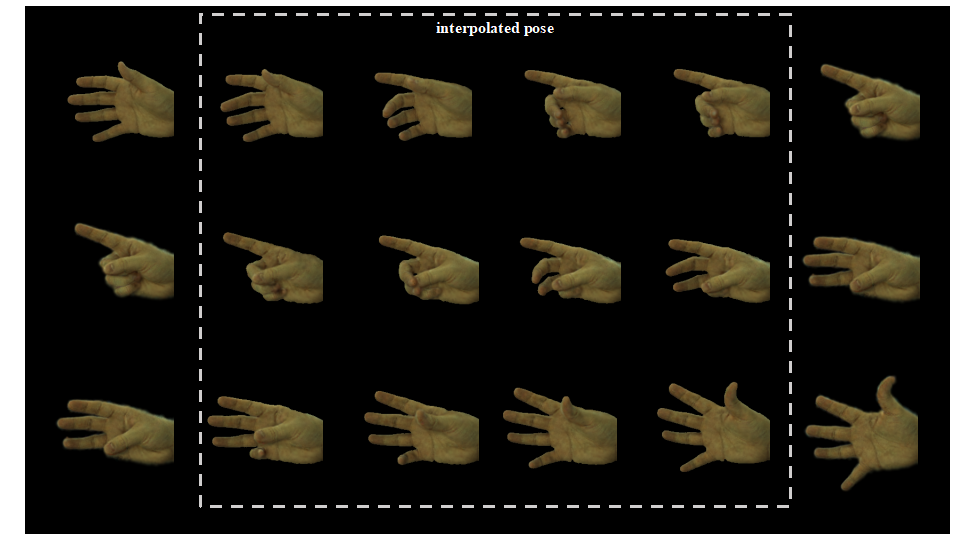}
	\caption{Qualitative results of the rendered images driven by interpolated poses. The first column (on the left) represents the initial pose, the last column (on the right) represents the ending pose, and the four columns in between illustrate the interpolated poses transitioning from the initial to the ending pose.}
	\label{fig:novel pose}
\end{figure*}

In \autoref{fig:novel view}, we present the rendering results of our method driven by three different poses from various viewpoints. Additionally, we randomly select three poses and interpolate them according to the rotation angles of the joints to generate interpolated poses. We then drive the modal to obtain the rendering results, as shown in \autoref{fig:novel pose}.

\section{Conclusion}
We have presented a novel 3DGS-based representation called JGHand which can reconstruct the human hand from RGB sequences and render photorealistic hand images in real. We proposed a zero-error transformation computation process for articulated hands. Furthermore, leveraging the advantages of 3D Gaussian explicit representation, we utilized the rendered depth images to simulate shadows generated by finger movement in real-time. Experimental results demonstrate that our method can render hand images that closely resemble the subject's hand and significantly reduce rendering time compared with related methods.

\textbf{Limitations and Future Work}.
Unlike other methods, the proposed JGHand does not require the parameters of the parameterized model but instead uses joint positions to drive the hand avatar. The computation of the transformation in our method is differentiable, enabling it to integrate with pose estimation networks for end-to-end training. This will greatly improve the method's generalization ability while optimizing the accuracy of pose estimation results. However, due to the triplane feature sampling strategy, our method requires training data that includes the complete texture of the hand. Otherwise, the texture of the missing parts cannot be inferred. In the future, we plan to explore one-shot methods to leverage the similarity of hand textures, conjecturing the complete hand texture from partial textures.

{\small
	\bibliographystyle{IEEEtran}
	\bibliography{egbib}
}

\appendices



\ifCLASSOPTIONcaptionsoff
  \newpage
\fi

\end{document}